\documentclass[a4]{article}
\usepackage[utf8]{inputenc} 
\usepackage{color}
\usepackage{graphicx}
\usepackage{amsmath}
\usepackage{url}
\usepackage{afterpage}
\usepackage{framed}
\date{\vspace{-5ex}}

\title{Kinematics and dynamics \\
of an egg-shaped robot with a gyro driven inertia actuator}

\author{
\normalsize
Norbert Michael Mayer\\
\scriptsize
Adaptive Embedded Systems and Robotics Laboratory (AES\&R),\\
\scriptsize
Dept. of Electrical Engineering and \\
\scriptsize
Advanced Institute of Manufacturing with High-tech Innovations (AIM-HI), \\
\scriptsize
National Chung Cheng University, Chia-Yi, \\
\scriptsize
Taiwan\\
\scriptsize
email: mikemayer@ccu.edu.tw}

\begin{document}
\maketitle

\begin{abstract}

The manuscript discusses still preliminary considerations with regard to the dynamics and kinematics of an egg shaped robot with an gyro driven inertia actuator. The method of calculation follows the idea that we would like to express the entire dynamic equations in terms of moments instead of forces. Also we avoid to derive the equations from a Lagrange function with constraints. The result of the calculations is meant to be applicable to two robot prototypes that have been build at the AES\&R Laboratory at the National Chung Cheng University in Taiwan. The design details of these prototypes are going to be outlined in \cite{egg2018}.

\end{abstract}

\vskip 0.5cm

\centerline{\includegraphics[height=6.0cm]{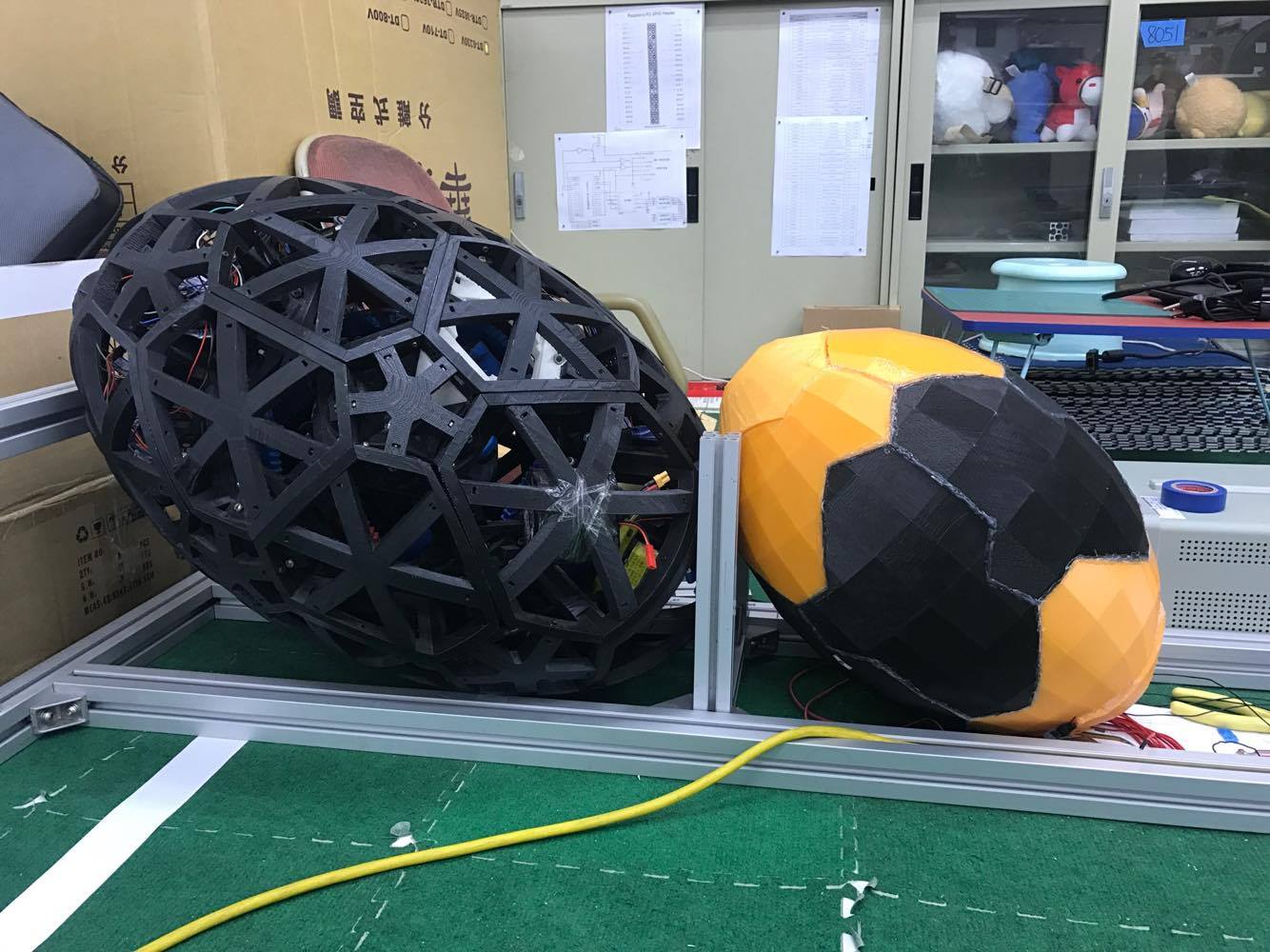}}

{\color{white} .}

\newpage
\small
\section*{Notations in formulas}

\begin{tabular}[t]{lll}
${\bf r}^l$							&$  = \{x^l, y^l, z^l\}$							& local coordinate of the robot's body \\ 
									&											& \\
$\phi$ 								&                								& angle of the outside gimbal relative to the robot's body \\ 
									&											& \\
$\psi$         						&	         									& angle of the inside gimbal relative to the outside gimbal\\
									&											& \\
$\alpha$, $\beta$, $\gamma$				&											& Euler angles used to describe a point \\
									&											& on the surface of the robots's body\\
$\alpha_v$, $\beta_v$, $\gamma_v$			&											& Euler angles towards \\
									&											& the vertical direction (towards the center of the earth)\\
$\beta_{p}$							&											& angle between robot axis and the direction towards the contact point \\
									&											& coordinates and the contact point between robot and ground\\
$\Theta_{com}$							&											& Total inertia tensor around the center of mass of the robot\\
$L$									&											& Total rotational impulse of the robot \\
$T_{mec}$								&											& Torque arising from motion of mechanical parts inside the robot  \\
$T_G$								& 											& Torque resulting from gravitation \\
$T_{fric}$							&											& Torque resulting from friction
\end{tabular}

\normalsize

\newpage

\section{Simulating the robot's dynamic behavior}

Dynamic behaviors usually are described in terms of forces, impulses and positions. 
Here, the aim here is to express all dynamic equations in terms of rotational impulses and torques, and orientations. In order to simulate the dynamic behavior a set of quantitative features of the egg have to be known:
\begin{itemize}
\item 1. $\Theta_{xy}$, 2. $\Theta_z$ are the components of the inertia tensor of the robot's body, 3. $M_R$ is the total mass of the robot.
\item 4. $R_l$ is the length of long axis of the robot, 5. $R_s$ are the lengths of the two short axes of the robot.
\item 6. $\Theta_\phi$ are the entries of the inertia in each direction for all parts that move along the outside ring of the gimbal. It is assumed that the mass distributions of all parts are rotational symmetric.
\item 7. $\Theta_\psi$ are the entries of the inertia tensor in each direction for all parts that move along the inside ring of the gimbal. Also, here the mass distribution is rotational symmetric.
\item 8. $\Theta_{rot}$ is component of the inertia tensor of the gyro along the rotation axis of the gyro.
\item 9. In addition, the twist-friction $\tau_{fcrit}$ has to be quantified, a value that indicates from what torque values the egg starts to twist against the ground.
\item 10. Stokes friction model $\rho_f$.
\end{itemize}

\section{Kinematics}

\subsection{Kinematic chain}

The total motion of the robot can be described in terms of a kinematic chain. One way to describe the kinematic chains is
\begin{itemize}
\item ground with a single global coordinate system,
\item contact point of the egg's body to the ground, that can be modeled similar to a ball joint (that is moving over the surface of the egg), 
state of the contact point is described by $\gamma$, $\beta$ and $\alpha$. This connection is passive, relates to outside forces, i.e. friction 
and gravitation.
\item the actuated hinge joint between the egg's body and the outer ring of the gimbal, i.e. that is $\phi$
\item the actuated hinge joint between the outer ring and inner ring, the hinge joint between the inside and outside ring, i.e. $\psi$.
\end{itemize}

\subsection{Coordinate systems}
The robot is a prolate rotational ellipsoid. As a convention the long axis of the ellipsoid is defined as the $z'$-axis of the local coordinate system of the egg, while $x'$ and $y'$ describe the other to Cartesian coordinates. The origin of the this local coordinate system is its center which is assumed to coincide with the center of mass.

In addition we define the global coordinate system. The vertical direction towards ground is $z$ while the two horizontal directions are $x$ and $y$. 

The relation between the local and global coordinate system can be expressed in terms of Euler angles. That is
rotations of body coordinates of the robot can be described by Euler angles in relation to the main symmetry axes of the robot body 
(i.e. the robots shell). 

In this paper the following definitions are used for the Euler angles: 
The angles $\alpha_v$, $\beta_v$, $\gamma_v$ describe subsequent rotations of the robot body in relation to the local $z'$-, $X$-axis (i.e. the new axis after the first rotation) , $z$-axis and result in a description of the robot's body in relation to the outside environment.
As a result of the definition the angle $\beta_v$ is the angle of the inclination of the egg with regard to the global vertical direction, i.e. the angle that is relevant for calculating the gravitational impact on the egg. 
The angles $\alpha_v$ and $\gamma_v$ are defined in that way that if both are equal to zero, the local (global, respectively) x-axes are identical to the inclination axis.

A second set of two Euler angles is used to describe the rotation of the gyro in relation to the robot's body. The Euler angles are called $\phi$, $\theta$ and $\psi$. The Euler angles correspond with the attitudes of rings of the gimbal that holds the gyro. So, $\phi$ is the angle between the outside ring of the gimbal in relation to the egg's body. $\psi$ is the angle between the outside and inside ring. Both angles are defined to be zero if the outside gimbal ring is in surface spanned by the local x and z axes.

\subsection{Rotation matrices}
We use Euler angles around $z$ and $x$ axes, and thus use the following active rotation matrices

\begin{eqnarray}
R_x(a) & = & 
\left( 
\begin{array}{ccc}
1 & 0 & 0 \\
0 & \cos a & -\sin a \\
0 &\sin a & \cos a  
\end{array}
\right) \nonumber \\
R_z(b) & = &
\left( 
\begin{array}{ccc}
\cos b & -\sin b & 0\\
\sin b & \cos b & 0\\
0&0 &1
\end{array}
\right).
\end{eqnarray}
Rotations from the global coordinate system to the local coordinate system are here active rotations, i.e.
\begin{equation}
\left( \begin{array} {c} x^l \\ y^l \\ z^l \end{array} \right) = R_z(\alpha) \cdot R_x(\beta) \cdot R_z(\gamma) \cdot  
\left( \begin{array} {c} x \\ y \\ z \end{array} \right).
\end{equation} 
The inverse transformation is controlled by passive rotations
\begin{equation}
\left( \begin{array} {c} x \\ y \\ z \end{array} \right) = R_z(-\gamma) \cdot R_x(-\beta) \cdot R_z(-\alpha) \cdot  
\left( \begin{array} {c} x^l \\ y^l \\ z^l \end{array} \right).
\end{equation}

\subsection{Robot surface in local coordinates}

Mathematically the robot has the shape of a prolate rotational ellipsoid. 
It is necessary to determine the coordinates of the the outer hull of the robot in relation to the center.

In the following the local coordinates of the surface of the robot's body are written as
\begin{eqnarray}
r^l(\beta, \alpha)=  |r(\beta)| \cdot R_x(-\beta) \cdot R_z(-\alpha) \cdot \left( \begin{array} {c} 1 \\ 0 \\ 0 \end{array} \right) =
|r(\beta)| \cdot
\left(
\begin{array}{l}
\cos \alpha \\
\sin \beta \cdot \sin \alpha \\
- \cos \beta \cdot \sin \alpha
\end{array}
\right)
,  
\end{eqnarray}
where the vector expresses the local coordinates $\left(x^l, y^l, z^l\right)$ of the shell in terms of Cartesian coordinates, $|r(\beta)|$ is according to the definitions of ellipsoids
\begin{equation}
|r(\beta)| = \frac{R_1 R_2} {\sqrt{R_2^2 \; \sin^2 \beta + R_1^2 \; \cos^2 \beta}}.
\end{equation}
$\beta$ is equivalent to a longitude and $\alpha$ is equivalent to an latitude angle if one considers both far ends of the egg as north- and south-pole. $R_1$ and $R_2$ are the radii of the shell, $R_1$ is assumed to be larger than $R_2$. 

For the following calculations it is also necessary to know 

\begin{equation}
r_\alpha(z^l) = \sqrt{R_2^2 -  R_2^2 \frac{(z^l)^2}{R_1^2} },
\end{equation}
that are the radii of the circles perpendicular to the long side of the egg, where $(0,0,z^l)$ is the local coordinate or the center point of the each circle. Alternatively, it is also possible to calculate, the same as a function of $\beta$:
\begin{equation}
r_\alpha (\beta) = \sqrt{R_2^2 - R_2^2 \frac{|r(\beta)|^2 cos^2 \beta}{R_1^2} }
\end{equation}

\begin{figure}
\centerline{
\includegraphics[height=5.0cm]{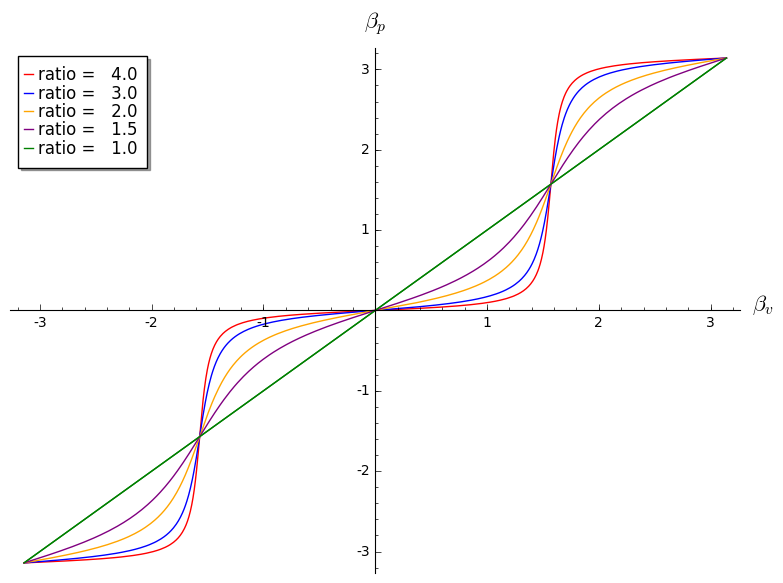}
\hskip 0.3cm
\includegraphics[height=5.0cm]{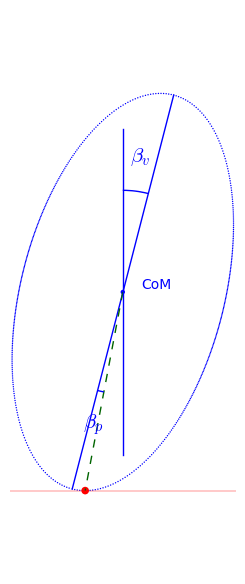}
}
\caption{ \label{incl_vs_cp_fig} Relation between contact point angle and inclination of the robot for different ratios between the long diameter and the short diameter.
 }
\end{figure}

\subsection{Contact point to the ground}
The contact point can be calculated as the maximum vertical point of the robot's body
\begin{equation}
-\cos \beta_v z^l + \sin \beta_v r_\alpha(z^l) = h
\end{equation}
Taking its derivative one gets
\begin{equation}
z^l_p = \frac{R_1 \cos \beta_v} {\sqrt{\cos^2 \beta_v + \frac{R_2^2}{R_1^2} \sin^2 \beta_v }}
\end{equation}
From here the relation 
\begin{equation}
\beta_p(\beta_v) = {\rm atan2} (r_\alpha(z^l_p), z^l_p ) 
\end{equation}
can be derived easily. Fig. \ref{incl_vs_cp_fig} depicts the resulting relation between the inclination angle of $\beta_v$ and the angle from the center towards 
the contact point.

\subsection{Motion kinematics}
Without slip and twist, the egg rolls like
\begin{eqnarray}
\dot{\gamma}_v &=& \dot{\alpha_v} \cos \beta_v \label{gammadyn_noslip}\\
\dot{x}_{p}  &=& |r(\beta_p)| \cos \gamma_v \cdot \dot{\beta_v} + r_\alpha (\beta_p) \dot{\alpha_v} \sin \gamma_v \\
\dot{y}_{p} &=& -|r(\beta_p)| \sin \gamma_v \cdot \dot{\beta_v} + r_\alpha (\beta_p) \dot{\alpha_v} \cos \gamma_v,  
\end{eqnarray}
where $\dot{x}_{p}$ and $\dot{y}_{p}$ refer to the motion of the contact point between the hull of the egg and the ground.
If one considers slip along $\dot{\gamma_v}$, one can modify eq. \ref{gammadyn_noslip} to
\begin{equation}
\dot{\gamma_v} = \dot{\alpha_v} \cos \beta_v + \dot{\gamma_v}_{slip} \label{gammadyn}.
\end{equation}

The motion of the center of the egg is:
\begin{equation}
\dot{\bf r} = 
\left(\begin{array}{c} 
\dot{x}_{p}\\
\dot{y}_{p}\\
0
\end{array}
\right)+ \frac{d}{dt} \left[
R_z(\gamma_v) \cdot R_X(\beta_p) 
\left( 
\begin{array}{c}
0 \\ 
|r(\beta_p)|\\
0
\end{array}
\right) \right]
\end{equation}

\subsection{Mass distribution inside the robot}
Within this work the robot is modeled is a dynamic systems that consists of 4 solid bodies, of which the attitudes are controlled by motor power. The robot is designed 
in the way that all parts' center of masses are positioned at the same position of the center of the robot. Table \ref{table_parts} depicts how the weight distributions are modeled. 

\subsection{Mechanics inside the robot and its kinematics}

The kinematic equations can be determined in the following way: 
On one hand if both motors go in the same direction the chain of gears that is placed on the inside side of the outside ring of the gimbal is blocked, thus the motion is directly transferred to the outside ring. On the hand if the motors rotate in the opposite direction the gears rotate while the outside ring does not move, and the inside ring is actuated. 
All other motions and postures can be created by a linear superposition of this two special cases which results in the following kinematic equation.
\begin{equation}
\left( \begin{array}{c} \phi \\ \psi \end{array} \right) = \left( \begin{array} {cc} 1 & 1 \\ 1 & -1 \end{array} \right) \cdot
\left( \begin{array}{c} \mu_1 \\ \mu_2 \end{array} \right),
\end{equation}
where $\mu_1$ and $\mu_2$ are the angles of the servo motors. The resulting angles $\alpha_u$ and $\beta_u$ indicate the position of the pole points of the axis of the gyro inside the actuator sphere.

\section{Dynamics of the robot}

In the following all dynamics are expressed in terms of moments around the contact point between the robot and the ground. We assume the contact point as one point of the robot's body that is not moving in this moment.

\subsection{Torque resulting from gravitation}
\begin{figure}
\centerline{
\includegraphics[height=8.0cm]{gravi1.png}
\includegraphics[height=8.0cm]{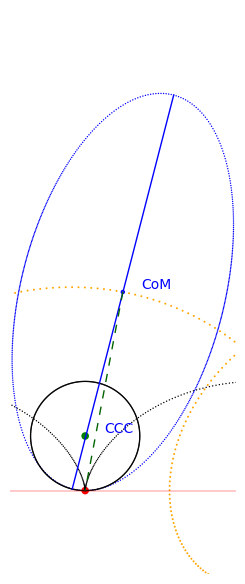}
\includegraphics[height=8.0cm]{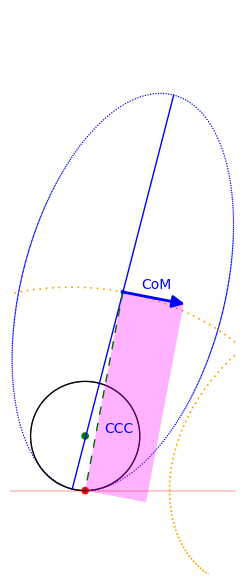}
}
\caption{ \label{gravitation_fig}
Graphical derivation of the impact of gravitation on the robot. Blue depicted is a potential shape of the egg. The red line is the horizontal ground line. The red dot represents the contact point. 
The curvature of the robot around the contact point defines a circle (in green). The motion trajectory of the robot around the contact point can be approximated by a rolling of the circle over the ground surface. 
Gravitation works vertically. The mechanical constraints redirect that force into the direction of the rolling motion. The simulation assumes all mass concentrated in the center (only with regard to the 
gravitation model). The resulting force direction is along a tangential line of cycloid. 
 }
\end{figure}

Construction of the direction of forces was tedious initially (cf. fig. \ref{gravitation_fig}). 
As a first step forces can be constructed along the rolling direction of the egg. 
The rolling of the egg can be described in terms of the curvature of the egg around the contact point. Rolling of all points on the rim of the circle 
can be described as cycloides. All points of a rolling disc can be constructed as curtate cycloids or prolate cycloids (if the the surface of the discs extends the defining circle). Here, the center of mass of the egg is modeled as rolling together with the curvature circle. 
Hence, it is a prolate or curtate cycloid. The force parallelogram results in the gravitational force component along the direction of the cycloid. 

In addition, the distance vector between the center of mass and the contact point. Then the resulting momentum is the cross vector product of the force component and the distance vector.

However, numerical simulations revealed that the following relations:
\begin{itemize}
\item The angle between the force vector and the direction vector between the center of mass and contact point is very close or identical to 90 degrees.
\item The prolate or curtate cycloid around the center of mass approximate a circle.
\end{itemize}
This simplifies the complexity of the calculation greatly. The absolute amount of the 

The gravitation force on the robot can then be calculated as 
\begin{equation}
{\cal F}_l = 9.81 \frac{m}{s^2} \cdot  \sin(\beta)
\end{equation}
and then the torque working on the robot is
\begin{equation}
|T_G| = r^l (\beta, \gamma) \cdot {\cal F}_l.
\end{equation}
The direction of the vector is perpendicular to the direction of the inclination, which is $\alpha$ from point of view of local coordinates.
Thus, the vector $T_l$ is 
\begin{equation}
T_G = |T_G| \cdot R_z(-\alpha) \cdot \left(\begin{array} {c} 1 \\ 0 \\ 0 \end{array} \right)
\end{equation}

\begin{table}[t]
\begin{tabular}[t]{|l|c|c|c|c|c|}
\hline
&&&&\\
n (frame)				& $R_n^r$ 	& $R_n^{r-}$	& $\omega^r_n	$  		&  $\hat{\Theta}_n$ & robot part \\
\hline
&&&\\
1 (shell)				& 			&	& $\omega		$ 		&
$\left( \begin{array}{ccc} \Theta_x && \\ & \Theta_x & \\ && \Theta_z \end{array} \right)$&
\raisebox{-.5\height}{\includegraphics[height=2cm, angle=0]{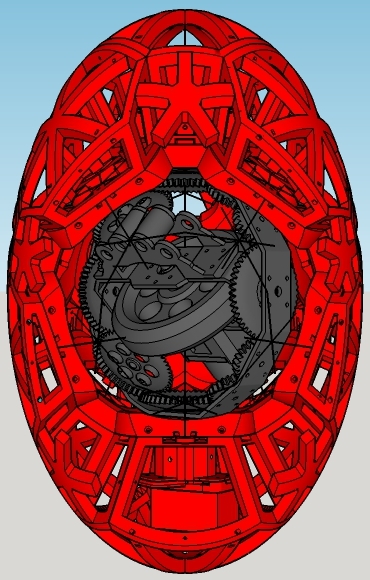}}\\[1cm]

2 (out. gimbal)		& $R_z(\phi)$	& $R_z(-\phi)$	& $\left(\begin{array}{c}0\\0\\1\end{array}\right) \cdot \dot{\phi}$ 	& 
$\left( \begin{array}{ccc} \Theta^\phi_x && \\ & \Theta^\phi_y & \\ && \Theta^\phi_x \end{array} \right)$&
\raisebox{-.5\height}{\includegraphics[height=2cm, angle=0]{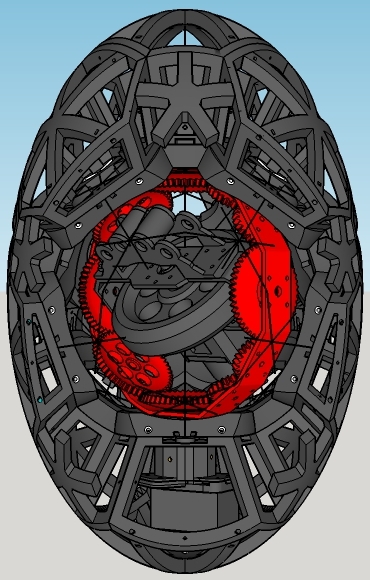}}
\\[1cm]

3 (in. gimbal) 		& $R_x(\psi)$	& $R_x(-\psi)$	& $\left(\begin{array}{c}1\\0\\0\end{array}\right) \cdot  \dot{\psi}$ 	& 
$\left( \begin{array}{ccc} \Theta^\psi_x && \\ & \Theta^\psi_x & \\ && \Theta^\psi_z \end{array} \right)$&
\raisebox{-.5\height}{\includegraphics[height=2cm, angle=0]{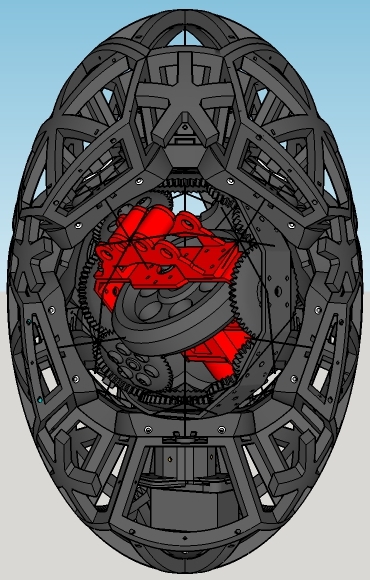}}\\[1cm]

4 (gyro)				& $R_z(\rho)$  & $R_z(-\rho)$	& $\left(\begin{array}{c}0\\0\\1\end{array}\right) \cdot \dot{\rho}$ 	& 
$\left( \begin{array}{ccc} \Theta^g_x && \\ & \Theta^g_x & \\ && \Theta^g_z \end{array} \right)$&
\raisebox{-.5\height}{\includegraphics[height=2.0cm, angle=0]{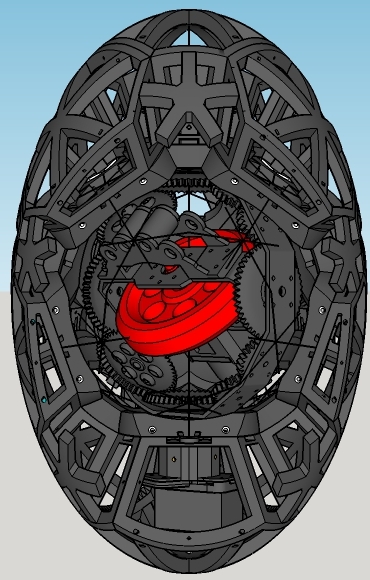}}\\[1cm]
\hline
\end{tabular}
\caption{A list of the parts of the robot that are considered as solid bodies for modeling the robot. The robot parts are symbolic, the shell is cut into half in order to allow a view into the inside. The rotation matrices $R^{r}_n$ indicate the relative rotation of the part versus the previous part of the kinematic chain, 
$R^{r-}_n$ its inverse. \label{table_parts}}
\end{table}

\afterpage{\clearpage}

\subsection{Rotational impulse of the motion system}

The dynamical effects of the robot's total motion system can be expressed in terms of a recursive sequence of equations. Each part of the robot has its own coordinate system. By design, all coordinate systems can be translated into each other by pure rotations. No translations are necessary. 

First the rotation vectors are calculated as the sum between the relative rotation vector between the $n$th part and the previous part and the total rotation of the 
previous part
\begin{equation}
\omega_n = \omega_n^r + R_n^r \cdot \omega_{n-1}.
\end{equation}
For $\omega_1$ one set $\omega$ which is the rotation vector. The corresponding rotation matrices $ R_n^r $ can be found in Table \ref{table_parts}.

At each level one can calculate the portion of the total rotational impulse in local coordinates of the current frame

\begin{table}[t]

\begin{framed}

\begin{eqnarray}
                        & L_4 = L_4^r = \hat{\Theta}_4 \cdot \omega_4 & \nonumber \\
         \nearrow       &                                             & \searrow \nonumber \\ 
  \omega_4 = \omega_4^r + R_4^r \cdot \omega_{3} &    &   L_{3} =R^{r-}_4 \cdot L_4 + \hat{\Theta}_3 \cdot \omega_3 \nonumber \\  
         \uparrow           \;\;\;\;\;\;\;       &    &    \;\;\;\;\;\;\;   \downarrow                                \nonumber \\
  \omega_3 = \omega_3^r + R_3^r \cdot \omega_{2} &    &   L_{2} = R^{r-}_3 \cdot L_3 + \hat{\Theta}_2 \cdot \omega_2 \nonumber \\  
         \uparrow           \;\;\;\;\;\;\;       &    &     \;\;\;\;\;\;\;  \downarrow                                \nonumber \\
  \omega_2 = \omega_2^r + R_2^r \cdot \omega     &    &   L = L_{1} = R^{r-}_2 \cdot L_2 + \hat{\Theta}_1 \cdot \omega_1 \nonumber 
\end{eqnarray}

\end{framed}
\caption{\label{tl1exp}
The recursive sequence of equations from which the total rotational impulse $L$ resulting from motions and actuations of 
the mechanical parts can be calculated. Values for all $\omega_n^r$, $R_n^r$, , $R_n^{r-}$. $\hat{\Theta}_n$ relate to the parts of the robot and can
be derived from Table \ref{table_parts}. The explicit form of $L$ is printed in App. \ref{l1exp}. }

\end{table}

\afterpage{\clearpage}

\begin{equation}
L_n^r = \hat{\Theta}_n \omega_n.
\end{equation}
From this the total rotational momentum can be calculated with

\begin{equation}
L_{n-1} =  R^{r-}_n \cdot L_n + L_{n-1}^r,
\end{equation}
where 

\begin{equation}
L_4 = L_4^r = \hat{\Theta}_4 \cdot \omega_4.
\end{equation}

The total angular momentum $L$ is equal to $L_1$.

As a next step one has to calculate the derivative of $L$ in time. Also the derivative can be expressed as a recursive formula

\begin{eqnarray}
\dot{\omega}_n &=& \dot{\omega}^r_n + \dot{R}_n^r \cdot \omega_{n-1} + R^r_n \cdot \dot{\omega}_{n-1}, \\
\dot{L}^r_n    &=& \hat{\Theta}_n \cdot \dot{\omega}_n,\\
\dot{L}_{n-1}  &=& \dot{R}^{r-}_n \cdot L_n + R^{r-}_n \cdot \dot{L}_n + \dot{L}^r_{n-1}.
\end{eqnarray}
Also, $\dot{\omega}_1 = \dot{\omega}$ and $\dot{ L_4}  = \dot{L_4^r}$. So, torque elicited from the mechanics of the robot can be written as
$T_{mec}=\dot{L}_1$. In order to prepare $T_{mec}$ in the appropriate form for the dynamic equations of the following section one has to find the 
linear factors $\hat{\Theta}_{com}$ and $B$ of 
\begin{equation}
T_{mec} = \hat{\Theta}_{com} \cdot \dot{\omega} + B, \label{tmec_eq}
\end{equation}
where the explicit form of $B$ can be found in App. \ref{Bexp}. $B$ does not depend on $\dot{\omega}$.
One gets for $\hat{\Theta}_{com}$:
\begin{equation}
\hat{\Theta}_{com} =
R_2^{r-} \left( R_3^{r-} \left( R_4^{r-} \hat{\Theta}_4 R_4^r + \hat{\Theta}_3 \right) R_3^r + \hat{\Theta}_2 \right) R_2^r + \hat{\Theta}_1.
\label{theta_com_1}
\end{equation}
If the robot is arranged in the way that $\Theta^\psi_x+\Theta^g_x=\Theta^\psi_z+\Theta^g_z$ then all inertia tensors become invariant against the corresponding rotations and thus eq. \ref{theta_com_1} can be simplified to
\begin{equation}
\hat{\Theta}_{com} = \hat{\Theta}_4 + \hat{\Theta}_3 + \hat{\Theta}_2 + \hat{\Theta}_1.
\end{equation}

\subsection{Virtual torque on a moving robot}
The Euler equations yield a rule how the rotating body perceives a virtual moment, that is
\begin{equation}
T_{cen} = \omega \times L
\end{equation}

\begin{table}[t]

\begin{framed}

\begin{eqnarray}
 \dot{\omega}_2 &=& \dot{\omega}^r_2 + \dot{R}_2^r \cdot \omega_{1} + R^r_2 \cdot \dot{\omega}_{1} \nonumber \\
 &\downarrow & \nonumber \\
 \dot{\omega}_3 &=& \dot{\omega}^r_3 + \dot{R}_3^r \cdot \omega_{2} + R^r_3 \cdot \dot{\omega}_{2} \nonumber \\
 &\downarrow & \nonumber \\
 \dot{\omega}_4 &=& \dot{\omega}^r_4 + \dot{R}_4^r \cdot \omega_{3} + R^r_4 \cdot \dot{\omega}_{3} \nonumber \\
 &\downarrow & \nonumber \\
 \dot{ L_4}     &=& \hat{\Theta}_n \cdot \dot{\omega}_4 \nonumber \\
 &\downarrow & \nonumber \\
 \dot{L}_{3}  &=& \dot{R}^{r-}_4 \cdot L_4 + R^{r-}_4 \cdot \dot{L}_4 + \hat{\Theta}_3 \cdot \dot{\omega}_3 \nonumber \\
 &\downarrow & \nonumber \\
 \dot{L}_{2}  &=& \dot{R}^{r-}_3 \cdot L_3 + R^{r-}_3 \cdot \dot{L}_3 + \hat{\Theta}_2 \cdot \dot{\omega}_2 \nonumber \\
 &\downarrow & \nonumber \\
 T_{mec} = 
 \dot{L}_{1}  &=& \dot{R}^{r-}_2 \cdot L_2 + R^{r-}_2 \cdot \dot{L}_2 + \hat{\Theta}_1 \cdot \dot{\omega}_1 \nonumber
\end{eqnarray}

\end{framed}
\caption{The recursive sequence of equations from which the partial rotational momentum $T_{mec}$ resulting from motions and actuations of 
the mechanical parts can be calculated. Values for all $\omega_n^r$, $R_n^r$, $\hat{\Theta}_n$ relate to the parts of the robot and can
be derived from Table \ref{table_parts}. Values for $\omega_n$ can be derived using the rules of Table \ref{tl1exp}.  Overall the depicted table shows 
equations that are derived from the derivative of each equation of Table \ref{tl1exp}.
The explicit form of $T_{mec}$ is printed in App. \ref{Bexp}. The explicit formula of the variable $B$ (cf. eq. \ref{tmec_eq}) can be derived by eliminating all 
summands that contain a factor of $\dot{\omega}_1$, i.e. setting $\dot{\omega}_1=0$.}

\end{table}

\subsection{Torque that is relevant for friction}
Twisting torque around the contact point between egg and ground surface can be expressed as 
\begin{equation}
t_{f} = \left( \begin{array}{c} 0 \\ 0 \\ 1 \end{array} \right) \cdot R_z (- \gamma) \cdot R_x(- \beta) \cdot R_z(- \alpha) \cdot
\left( T_{G} + T_{mec} + T_{cen} \right),
\end{equation}
where $t_f$ is the friction coefficient. There are two possible cases: 
\begin{itemize}
\item Static friction (stiction), where the egg does not twist and slip around its axis.
\item Friction, that is modeled according to Stokes
\end{itemize}
The transition between these two situations is characterized by a critical value for $t_f$, in the following called $\tau_{fcrit}$.
In the stiction case, i.e. $t_f \leq \tau_{fcrit}$:
\begin{equation}
T_{frict} = - \left( \begin{array} {c} 0 \\ 0 \\ 1 \end{array} \right) \cdot t_{f}.
\end{equation}
The Stokes case (i.e. $t_f > \tau_{fcrit}$) can be described by
\begin{equation}
T_{frict} = -  \left( \begin{array} {c} 0 \\ 0 \\ 1 \end{array} \right) \cdot \rho_f \cdot \dot{\gamma}_{slip}.
\end{equation}
Since the torque here is expressed in global terms it necessary to transfer it into the local coordinate system
\begin{equation}
T^l_{frict} = R_z( \alpha) \cdot  R_x( \beta) \cdot R_z ( \gamma) \cdot T_{frict}.
\end{equation}
In the scope of this work we do not consider any other types of slipping, that is we do not consider that the robot could slide along the x and y directions.
Thus, the contact point between

\subsection{Dynamic equations}
The following considerations base on the fundamental dynamic equation of a rotating rigid body around its center of mass
\begin{equation}
\dot{L} = M, \label{dyn_1}
\end{equation}
i.e. the derivative of the rotational impulse is equal to the momentum that works on the body from the outside world, thus
\begin{equation}
M = T^l_{G} + T^l_{frict} + T_{cen}. \label{dyn_2}
\end{equation}
For $\dot{L}$ we have
\begin{equation}
\dot{L} = T_{mec} = \hat{\Theta}_{com} \cdot \dot{\omega} + B \label{dyn_3}
\end{equation}
Setting eq. \ref{dyn_2} and \ref{dyn_3} in eq. \ref{dyn_1} results in
\begin{equation}
 \hat{\Theta}_{com} \cdot \dot{\omega} + B =  T^l_{G} + T^l_{frict} + T_{cen}.
\end{equation} 
and thus one can get 
\begin{eqnarray}
\dot{\omega} &=& \hat{\Theta}^{-1}_{com} \left( -B + T^l_{G} + T^l_{frict}  + T_{cen} \right)\\
\left( \begin{array}{c} \dot{\alpha} \\ \dot{\beta} \\ \dot{\gamma} \end{array} \right) &=& \left( 
\begin{array} {ccc}  
\sin \beta \sin \gamma & \sin \beta \cos \gamma & \cos \beta     \\
 \cos \gamma  & -\sin \gamma &     \\
            &              & 1 
\end{array}
\right) \cdot \omega,
\end{eqnarray}
which represents an explicit differential equation and a complete set of dynamic equations of the robot. 

\newpage
\appendix

\section{Explicit formula of $L$ \label{l1exp}}

The explicit formulas for $L$ and $\dot{L}$ have been compiled using a small python script:

\begin{eqnarray}
L_1&=& 
 \hat{\Theta}_1  \cdot  \omega  + 
\\  \nonumber & &  R_2^{r-}  \cdot  R_3^{r-}  \cdot  R_4^{r-} 
\cdot  \hat{\Theta}_4  \cdot  \omega^r_4  + 
 R_2^{r-}  \cdot  R_3^{r-}  \cdot  \hat{\Theta}_3  \cdot  \omega^r_3  + 
\\  \nonumber & &  R_2^{r-}  \cdot  R_3^{r-}  \cdot  R_4^{r-} 
\cdot  \hat{\Theta}_4  \cdot  R_4^r  \cdot  \omega^r_3  + 
 R_2^{r-}  \cdot  \hat{\Theta}_2  \cdot  \omega^r_2  + 
\\  \nonumber & &  R_2^{r-}  \cdot  \hat{\Theta}_2  \cdot  R_2^r
\cdot  \omega  + 
 R_2^{r-}  \cdot  R_3^{r-}  \cdot  \hat{\Theta}_3  \cdot  R_3^r  \cdot 
\omega^r_2  + 
\\  \nonumber & &  R_2^{r-}  \cdot  R_3^{r-}  \cdot 
\hat{\Theta}_3  \cdot  R_3^r  \cdot  R_2^r  \cdot  \omega  + 
\\  \nonumber & &  R_2^{r-}  \cdot  R_3^{r-}  \cdot  R_4^{r-} 
\cdot  \hat{\Theta}_4  \cdot  R_4^r  \cdot  R_3^r  \cdot  \omega^r_2  + 
\\  \nonumber & &  R_2^{r-}  \cdot  R_3^{r-}  \cdot  R_4^{r-} 
\cdot  \hat{\Theta}_4  \cdot  R_4^r  \cdot  R_3^r  \cdot  R_2^r  \cdot 
\omega 
 \end{eqnarray}

\newpage
\section{Explicit formula of $T_{mec}$ \label{Bexp}}

Just as a reminder the momentum resulting from mechanical processes inside the egg is
\begin{equation}
T_{mec} = \hat{\Theta}_{com} \cdot \dot{\omega} + B, \nonumber
\end{equation}
where 
\begin{equation}
\hat{\Theta}_{com} = \hat{\Theta}_4 + \hat{\Theta}_3 + \hat{\Theta}_2 + \hat{\Theta}_1. \nonumber
\end{equation}

The value of $B$ is
\tiny
\begin{eqnarray} 
B&=& 
 R_2^{r-}  \cdot  R_3^{r-}  \cdot  R_4^{r-}  \cdot  \hat{\Theta}_4 
\cdot  \dot{\omega}^r_4  + 
\\  \nonumber & &  R_2^{r-}  \cdot  R_3^{r-}  \cdot 
\hat{\Theta}_3  \cdot  \dot{\omega}^r_3  + 
\\  \nonumber & &  R_2^{r-}  \cdot  R_3^{r-}  \cdot  R_4^{r-} 
\cdot  \hat{\Theta}_4  \cdot  R_4^r  \cdot  \dot{\omega}^r_3  + 
 R_2^{r-}  \cdot  \hat{\Theta}_2  \cdot  \dot{\omega}^r_2  + 
\\  \nonumber & &  R_2^{r-}  \cdot  \hat{\Theta}_2  \cdot 
\dot{R}_2^r  \cdot  \omega  + 
 R_2^{r-}  \cdot  R_3^{r-}  \cdot  \hat{\Theta}_3  \cdot  R_3^r  \cdot 
\dot{\omega}^r_2  + 
\\  \nonumber & &  R_2^{r-}  \cdot  R_3^{r-}  \cdot 
\hat{\Theta}_3  \cdot  R_3^r  \cdot  \dot{R}_2^r  \cdot  \omega  + 
\\  \nonumber & &  R_2^{r-}  \cdot  R_3^{r-}  \cdot  R_4^{r-} 
\cdot  \hat{\Theta}_4  \cdot  R_4^r  \cdot  R_3^r  \cdot 
\dot{\omega}^r_2  + 
\\  \nonumber & &  R_2^{r-}  \cdot  R_3^{r-}  \cdot  R_4^{r-} 
\cdot  \hat{\Theta}_4  \cdot  R_4^r  \cdot  R_3^r  \cdot  \dot{R}_2^r 
\cdot  \omega  + 
 R_2^{r-}  \cdot  R_3^{r-}  \cdot  \dot{R}_4^{r-}  \cdot  \hat{\Theta}_4
\cdot  \omega^r_4  + 
\\  \nonumber & &  \dot{R}_2^{r-}  \cdot  R_3^{r-}  \cdot 
R_4^{r-}  \cdot  \hat{\Theta}_4  \cdot  \omega^r_4  + 
 R_2^{r-}  \cdot  \dot{R}_3^{r-}  \cdot  R_4^{r-}  \cdot  \hat{\Theta}_4
\cdot  \omega^r_4  + 
\\  \nonumber & &  R_2^{r-}  \cdot  \dot{R}_3^{r-}  \cdot 
\hat{\Theta}_3  \cdot  \omega^r_3  + 
\\  \nonumber & &  R_2^{r-}  \cdot  R_3^{r-}  \cdot  R_4^{r-} 
\cdot  \hat{\Theta}_4  \cdot  \dot{R}_4^r  \cdot  \omega^r_3  + 
\\  \nonumber & &  R_2^{r-}  \cdot  R_3^{r-}  \cdot 
\dot{R}_4^{r-}  \cdot  \hat{\Theta}_4  \cdot  R_4^r  \cdot  \omega^r_3 
+ 
\\  \nonumber & &  \dot{R}_2^{r-}  \cdot  R_3^{r-}  \cdot 
R_4^{r-}  \cdot  \hat{\Theta}_4  \cdot  R_4^r  \cdot  \omega^r_3  + 
\\  \nonumber & &  R_2^{r-}  \cdot  \dot{R}_3^{r-}  \cdot 
R_4^{r-}  \cdot  \hat{\Theta}_4  \cdot  R_4^r  \cdot  \omega^r_3  + 
 \dot{R}_2^{r-}  \cdot  R_3^{r-}  \cdot  \hat{\Theta}_3  \cdot 
\omega^r_3  + 
\\  \nonumber & &  \dot{R}_2^{r-}  \cdot  \hat{\Theta}_2  \cdot 
\omega^r_2  + 
 \dot{R}_2^{r-}  \cdot  \hat{\Theta}_2  \cdot  R_2^r  \cdot  \omega  + 
\\  \nonumber & &  R_2^{r-}  \cdot  R_3^{r-}  \cdot 
\hat{\Theta}_3  \cdot  \dot{R}_3^r  \cdot  \omega^r_2  + 
\\  \nonumber & &  R_2^{r-}  \cdot  R_3^{r-}  \cdot 
\hat{\Theta}_3  \cdot  \dot{R}_3^r  \cdot  R_2^r  \cdot  \omega  + 
\\  \nonumber & &  R_2^{r-}  \cdot  R_3^{r-}  \cdot  R_4^{r-} 
\cdot  \hat{\Theta}_4  \cdot  R_4^r  \cdot  \dot{R}_3^r  \cdot 
\omega^r_2  + 
\\  \nonumber & &  R_2^{r-}  \cdot  R_3^{r-}  \cdot  R_4^{r-} 
\cdot  \hat{\Theta}_4  \cdot  R_4^r  \cdot  \dot{R}_3^r  \cdot  R_2^r 
\cdot  \omega  + 
 R_2^{r-}  \cdot  \dot{R}_3^{r-}  \cdot  \hat{\Theta}_3  \cdot  R_3^r 
\cdot  \omega^r_2  + 
\\  \nonumber & &  R_2^{r-}  \cdot  \dot{R}_3^{r-}  \cdot 
\hat{\Theta}_3  \cdot  R_3^r  \cdot  R_2^r  \cdot  \omega  + 
\\  \nonumber & &  R_2^{r-}  \cdot  R_3^{r-}  \cdot  R_4^{r-} 
\cdot  \hat{\Theta}_4  \cdot  \dot{R}_4^r  \cdot  R_3^r  \cdot 
\omega^r_2  + 
\\  \nonumber & &  R_2^{r-}  \cdot  R_3^{r-}  \cdot  R_4^{r-} 
\cdot  \hat{\Theta}_4  \cdot  \dot{R}_4^r  \cdot  R_3^r  \cdot  R_2^r 
\cdot  \omega  + 
\\  \nonumber & &  R_2^{r-}  \cdot  R_3^{r-}  \cdot 
\dot{R}_4^{r-}  \cdot  \hat{\Theta}_4  \cdot  R_4^r  \cdot  R_3^r  \cdot
\omega^r_2  + 
\\  \nonumber & &  R_2^{r-}  \cdot  R_3^{r-}  \cdot 
\dot{R}_4^{r-}  \cdot  \hat{\Theta}_4  \cdot  R_4^r  \cdot  R_3^r  \cdot
R_2^r  \cdot  \omega  + 
\\  \nonumber & &  \dot{R}_2^{r-}  \cdot  R_3^{r-}  \cdot 
R_4^{r-}  \cdot  \hat{\Theta}_4  \cdot  R_4^r  \cdot  R_3^r  \cdot 
\omega^r_2  + 
\\  \nonumber & &  \dot{R}_2^{r-}  \cdot  R_3^{r-}  \cdot 
R_4^{r-}  \cdot  \hat{\Theta}_4  \cdot  R_4^r  \cdot  R_3^r  \cdot 
R_2^r  \cdot  \omega  + 
\\  \nonumber & &  R_2^{r-}  \cdot  \dot{R}_3^{r-}  \cdot 
R_4^{r-}  \cdot  \hat{\Theta}_4  \cdot  R_4^r  \cdot  R_3^r  \cdot 
\omega^r_2  + 
\\  \nonumber & &  R_2^{r-}  \cdot  \dot{R}_3^{r-}  \cdot 
R_4^{r-}  \cdot  \hat{\Theta}_4  \cdot  R_4^r  \cdot  R_3^r  \cdot 
R_2^r  \cdot  \omega  + 
 \dot{R}_2^{r-}  \cdot  R_3^{r-}  \cdot  \hat{\Theta}_3  \cdot  R_3^r 
\cdot  \omega^r_2  + 
\\  \nonumber & &  \dot{R}_2^{r-}  \cdot  R_3^{r-}  \cdot 
\hat{\Theta}_3  \cdot  R_3^r  \cdot  R_2^r  \cdot  \omega 
 \end{eqnarray}

\end{document}